\documentclass[10pt,twocolumn,letterpaper]{article}

\usepackage{iccv}
\usepackage{times}
\usepackage{epsfig}
\usepackage{graphicx}
\usepackage{amsmath}
\usepackage{amssymb}
\usepackage{multirow}
\usepackage{xcolor}
\usepackage{comment}
\usepackage{subcaption}
\usepackage{float}
\usepackage[hang,flushmargin]{footmisc}

\usepackage[breaklinks=true,bookmarks=false]{hyperref}

\iccvfinalcopy % *** Uncomment this line for the final submission

\def\iccvPaperID{****} % *** Enter the CVPR Paper ID here

\begin{document}

%%%%%%%%% TITLE
\title{\textit{LEAF-QA}: Locate, Encode \& Attend for Figure Question Answering}

\author{Ritwick Chaudhry\\
Carnegie Mellon University\\
{\tt\small rchaudhr@andrew.cmu.edu}
% To save space, use !either the email address or home page, not both
\and
Sumit Shekhar\\
Adobe Research\\
{\tt\small sushekha@adobe.com}
\and
Utkarsh Gupta\\
IIT Roorkee\\
{\tt\small utkarshgupta@ec.iitr.ac.in}
\and 
Pranav Maneriker\\
Ohio State University\\
{\tt\small maneriker.1@osu.edu}
\and
Prann Bansal\\
IIT Kanpur\\
{\tt\small prann@iitk.ac.in}
\and
Ajay Joshi\\
IIT Madras\\
{\tt\small cs15b047@smail.iitm.ac.in}
}

\maketitle

%%%%%%%%% ABSTRACT
\begin{abstract}
 We introduce \textit{LEAF-QA}, a comprehensive dataset of $250,000$ densely annotated  figures/charts, constructed from real-world open data sources, along with $~2$ million question-answer (QA) pairs querying the structure and semantics of these charts. \textit{LEAF-QA} highlights the problem of multimodal QA, which is notably different from conventional visual QA (VQA), and has recently gained interest in the community. Furthermore, \textit{LEAF-QA} is significantly more complex than previous attempts at chart QA, \textit{viz.} FigureQA and DVQA, which present only limited variations in chart data. \textit{LEAF-QA} being constructed from real-world sources, requires a novel architecture to enable question answering. To this end, \textit{LEAF-Net}, a deep architecture involving chart element localization, question and answer encoding in terms of chart elements, and an attention network is proposed. Different experiments are conducted to demonstrate the challenges of QA on \textit{LEAF-QA}. The proposed architecture, \textit{LEAF-Net} also considerably advances the current state-of-the-art on FigureQA and DVQA.
\end{abstract}

\begin{figure*}[h]
    \centering
    \includegraphics[width=0.95\linewidth]{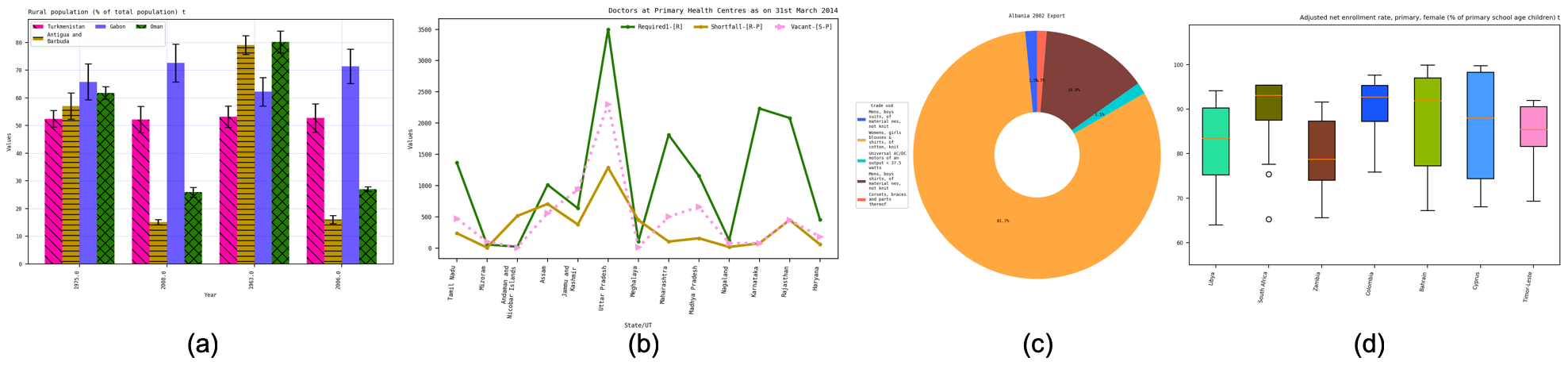}
    \caption{Sample chart images from \textit{LEAF-QA} corpus for (a) bar plot, (b) a line graph, (c) pie/donut and (d) box plot. Each image represents data from a real world source, thus making question answering a challenging task.}
    \label{fig:LEAF-QA_Samples}
\end{figure*}

%%%%%%%%% BODY TEXT
\section{Introduction}
Charts/figures (used interchangeably hereafter) are effective and widely used representations of data in documents in almost every domain. Thus, for understanding of documents by machines, it is critical to investigate into robust technologies that can interpret charts and figures, and have the ability to answer varied user queries. With advances in deep learning technologies, it is possible to interpret images such as charts, provided a large annotated corpus is available. However, it is a challenging problem to gather large scale data consisting of charts along with annotations mainly because (a) legal and copyright issues crop up while extracting charts from different documents, (b) scraping the charts from documents doesn't provide relevant annotations of the chart elements, as the original data values are lost when the chart for the data is produced.

Current datasets on figure question answering such as FigureQA~\cite{ebrahimi2018figureqa} and DVQA~\cite{kafle2018dvqa} have charts generated from synthetic data which have limited ranges of data values. Further, while FigureQA has questions having only binary Yes/No answers, the answer vocabulary of DVQA has only around thousand unique answers. Such assumptions do not hold in case of charts in the real-world, where each chart can potentially have a distinct vocabulary, with words as well as data values not seen before. Furthermore, FigureQA~\cite{ebrahimi2018figureqa} chart labels are set to the color name of the corresponding plot element in the chart (\eg a purple line would have the legend label `purple'). This may cause question answering models to associate colors instead of semantics of the charts with the answers. DVQA~\cite{kafle2018dvqa} introduces diversity through randomizing legend labels, colors and axis labels, however it is limited to only bar charts. Moreover, as synthetic data is used in DVQA generation, there is a limited variation in title, axis and legend labels, \eg charts in DVQA do not have any axis labels more than six characters long. FigureSeer \cite{siegel2016figureseer} presents a dataset of around $60,000$ figures scraped from different research papers, but a set of only $600$ figures are densely annotated using Mechanical Turk. Further, a set of heuristics are presented to extract various chart elements. However, this may not generalize well to different charts, because of considerable variations in the arrangement of chart elements. Moreover, previous datasets have fixed question templates which don't take into consideration the possible variations in natural user queries.

To this end, \textit{LEAF-QA} provides a large-scale corpus of a variety of charts, which mitigates the pressing problems in existing chart question answering datasets. Firstly, charts are extracted from varied real-world data sources, like the government census or financial data, thus avoiding biases of synthesized data, as well as providing much more complex variations than the previous attempts. Chart images are plotted as multiple visualization categories \textit{viz.} bar (stacked and grouped) graphs, pie/donut charts, box plots, scatter plots and line graphs. The charts are complemented with dense annotations of the different chart elements, with bounding boxes or masks as relevant (\textit{e.g.} for pie wedges). Different question-answer pairs querying the chart images are provided, covering both query types: structural and relational. Moreover, questions are randomly selected from para-phrasings of question templates, to prevent question answering models from memorizing templates. \textit{LEAF-QA} consists of a corpus of $~200,000$ training chart images and $~1.6$ million question-answer pairs. For testing, $40,000$ chart images and $~0.4$ million question-answer pairs are provided. Additionally, a novel test set generated from unseen data sources is also provided to test the generalization of learning frameworks.

\textit{LEAF-QA} has a large number of unique answers, which include strings contained within the charts. Hence, current deep learning architectures for VQA with fixed answer dictionaries can not be directly applied, as the test set contains answers which are not present in the training corpus. A novel framework, \textit{LEAF-Net} using detection and parsing of chart elements, followed by encoding questions and answers in terms of the detected chart elements is proposed to handle this. A comprehensive evaluation of the detection of chart elements as well as the performance of the network on different question types is detailed. A thorough comparison of \textit{LEAF-Net} on architectures proposed in DVQA and FigureQA is further presented.

\section{Related Work} \label{sec:RelatedWork}
This section describes prior art related to question answering on charts. 

\textbf{Figure Understanding:} There has been recent interest in analyzing figures and charts, particularly to understand the type of visualization and for data extraction from the chart images. Savva et al.~\cite{savva2011revision} describe algorithms to extract data from pie and bar charts, particularly to re-visualize them. Further, interactive methods for bar chart extraction have been studied~\cite{jung2017chartsense, rohatgi2011webplotdigitizer}. Cliche et al.~\cite{cliche2017scatteract} describe an object detection framework for extracting scatter plot elements. Similarly, an analysis for line plot extraction has been presented by Nair et al.~\cite{nair2015automated}. There have also been attempts at indexing of figures \cite{ray2015architecture, battle2018beagle} for search and classification. B{\"o}schen et al.~\cite{boschen2015multi} and Poco et al.~\cite{poco2017reverse} describe methods for improving text and symbol extraction from figures. Harper et al.~\cite{harper2014deconstructing} describe a framework to restyle different kinds of visualizations, through maneuvering the data in the SVGs.

\textbf{Visual Question Answering:} Learning to answer questions based on natural images has been an area of extensive research in recent years. Several datasets including DAQUAR~\cite{malinowski2014multi}, COCO-QA~\cite{ren2015exploring}, VQA~\cite{antol2015vqa}, Visual7w~\cite{zhu2016visual7w} and MovieQA~\cite{Tapaswi_2016_CVPR} have been proposed to explore different facets of question answering on natural images and videos. Correspondingly, methods using attention~\cite{xu2015show, yang2016stacked, Kazemi2017}, neural modules~\cite{andreas2016neural} and compositional modeling~\cite{fukui2016multimodal} have been explored. There has been related work on question answering on synthetic data ~\cite{johnson2017clevr, Johnson_2017_ICCV}. However, the current work is most related to recent work on multimodal question answering~\cite{singh2019towards, biten2019icdar},  which show that current VQA do not perform well while reasoning on text in natural images, and hence, there is a need to learn image and scene text jointly for question/answering. 

\textbf{Object Detection:} There have been significant advances in object detection frameworks, especially networks based on the recent region-based convolutional neural networks (R-CNNs)~\cite{girshick2015fast,he2017mask,ren2015faster}, YOLO~\cite{redmon2016you} and SSD~\cite{liu2016ssd}. As the elements in \textit{LEAF-QA} charts are both rectangular boxes as well as elements like pie wedges, the Mask R-CNN~\cite{he2017mask} is used for parsing chart elements.

\section{\textit{LEAF-QA} dataset} \label{sec:DatasetSec}
In this section, the generation of figures and question-answer pairs in \textit{LEAF-QA} is explained, along with details of the training and test corpora.

\subsection{Data Sources}
\textit{LEAF-QA} uses public data tables available from various online sources, which are listed below: 

   1. World Development Indicators (World Bank) \footnote{www.datacatalog.worldbank.org/dataset/world-development-indicators}
    2. Gender Statistics (World Bank) \footnote{www.datacatalog.worldbank.org/dataset/gender-statistics}
    3. Government of India Open Data \footnote{www.visualize.data.gov.in}
    4. Commodity Trade Statistics (UN data) \footnote{www.kaggle.com/unitednations/global-commodity-trade-statistics/data}
    5. US Census Data (for 2000 and 2010) \footnote{www.kaggle.com/muonneutrino/us-census-demographic-data/data}
    6. Price Volume Data for Stocks and ETFs \footnote{www.kaggle.com/borismarjanovic/price-volume-data-for-all-us-stocks-etfs}

\begin{table*}[ht!]
\centering
\small
\captionsetup{skip=0pt}
\begin{tabular}{c|cc|cc|c|c|cc|c|c}
\hline
\multirow{2}{*}{\textbf{Corpus}} & \multicolumn{2}{c|}{\textbf{Group Bar}} & \multicolumn{2}{|c|}{\textbf{Stack bar}} & \textbf{Pie} & \textbf{Donut} & \multicolumn{2}{c|}{\textbf{Box}} & \textbf{Line} & \textbf{Scatter} \\
& Horz. & Vert. & Horz. & Vert. & & & Horz. & Vert. & & \\
\hline\hline
Training & 8,588 & 8,562 & 13,726 & 13,414 & 7,008 & 21,206 & 21,012 & 20,961 & 41,935 & 41,710 \\
Testing & 1,674 & 1,757 & 2,690 & 2,736 & 1,445 & 4,221 & 4,167 & 4,203 & 8,392 & 8,336 \\
Novel Test & 626 & 624 & 639 & 611 & 1250 & 1250 & 52 & 58 & 2,500 & 550 \\
\hline
\end{tabular}
\begin{tabular}{c|cc|cccc}
\hline 
\multirow{2}{*}{\textbf{Corpus}} & \multicolumn{2}{c|}{\textbf{Questions}} & \multicolumn{3}{c}{\textbf{Answers}} \\
& Relational & Structural & Chart Vocabulary & Common Vocabulary & Chart Type\\
\hline\hline 
Training & 13,30,395 & 1,98,904 & 10,06,229 & 3,59,498 & 1,63,572 \\
Testing & 2,66,318 & 39,764 & 2,01,469 & 71,929 & 32,684\\
Novel Test & 60,697 & 10,408 & 47,706 & 17,289 & 6,110\\
\hline
\end{tabular}
\caption{Statistics of different chart types in \textit{LEAF-QA}.}
\label{tab:CorpusBreakUp}
\end{table*}

\subsection{Data pre-processing}
Real-world data tables are pre-processed to make them suitable for creating different types of charts and tackle the problem of missing data. The steps are as follows:

\textbf{Merging data}: Data series sharing the label space, like the x and y-label names are combined to get an integrated table. For example, from stocks database consisting of multiple tables reflecting stock prices for different companies, a single table containing information for all the companies is created. This bigger table can hence be used to make charts that compare stock prices of different companies.

\textbf{Decomposing Tables}: Tables representing data series of multiple indicators are split along several axes to create a pool of smaller tables. The split can be done on legend labels, x-labels or y-labels for generating varied charts. For example, the ``world development indicators (World Bank)'' is a table assessing various indicators of different countries for multiple years. This table is decomposed into smaller tables \textit{e.g.,} one representing an indicator for one country over time and another an indicator of different countries for a given year.

\textbf{Handling Missing/Noisy Data}: Symbols such as N/A in columns and other missing values are imputed with a random value drawn from the normal distribution fitted to data in the column. Rows containing data aggregated from multiple rows, such as the sum of columns, are removed. Columns with complex alpha-numeric data like serial numbers or hexadecimal hashing are eliminated.

\subsection{Figure/Chart Generation}
From each table, multiple charts are created, based on a selection of different sets of columns, followed by a randomized selection of rows from which data is to be plotted. The chart generation is done using the Matplotlib library\footnote{www.matplotlib.org/}.
The different chart types are created as follows:
\begin{itemize}
\setlength{\itemsep}{0.5pt}
    \item Pie charts and donut charts are created from columns having numerical data in the table with their categorical label information being present as row headers.
    \item Vertical/horizontal bar, stacked bar and grouped bar graphs are created using multiple numerical columns from a table with the categorical label information taken from row headers.
    \item For vertical/horizontal boxplots, statistics such as median, maximum, minimum, standard deviation are calculated for columns having numeric data.
    \item Line graphs are developed using multiple rows of the table (for example, time series data).
    \item Scatter plots are constructed using two columns having numeric data.
\end{itemize}

\subsection{Figure/Chart Style Variations}
The following steps are used to make charts more readable and introduce variations in chart elements to emulate real-world data:

\textbf{Legend color palette creation}: Colors of plot elements are made distinct from each other by using pre-defined color palettes having optimally distinct colors, made using an online tool\footnote{http://tools.medialab.sciences-po.fr/iwanthue/index.php}.

\textbf{Skew/overlap removal}: In order to prevent skewed plots, the columns which are chosen to be plotted together have values with similar order of magnitude.

\textbf{Element variations}: Variations are introduced in chart components to emulate varied positions/fonts/sizes as is observed in real-world data. The following features are varied: (i) Title positioning, (ii) Font Families and sizes, (iii) Marker style and line width and style, (iv) Grid style and colors, (v) Legend placement, border and position of the legend entries, (vi) Width of bars, (vii) Inner and outer radii of pies, (viii) Presence of error bars.

\subsection{Annotation Generation}
\textbf{Bounding Box Generation}: Firstly, tight bounding boxes are extracted for chart elements like axis labels, title, legend box, legend items and plot elements, using Matplotlib functions, and corresponding labels are assigned. The plot elements from different chart types like bars (horizontal and vertical, stacked), pies, scatter plot markers, lines and boxes are each defined as different classes, leading to a total of $20$ classes of chart elements.

\textbf{Mask Generation}: For some plot elements, bounding boxes are not representative of the actual ground truth. So masks are generated for these elements:
\begin{itemize}
\setlength{\itemsep}{0.5pt}
    \item \textbf{Pie/Donut Plots}: Firstly, radius and center of the pie chart are estimated using the positions of the bounding boxes. This is then used to create a mesh representing each pie element. To approximate the circle, a polygon with vertices at every 1 degree is constructed. 
    \item \textbf{Line Plots}: A single bounding box is not sufficient to represent trends in the line plot. Hence, multiple bounding boxes, each between two consecutive x-labels, are used to represent a line plot.
\end{itemize}

\subsection{Question-Answer Generation}
Questions in \textit{LEAF-QA} have been based upon the analytical reasoning questions in the Graduate Record Examination (GRE)\footnote{https://magoosh.com/gre/2016/gre-data-interpretation-practice/}, hence reflecting real-world situations. Furthermore, questions for each type of graph vary based on the suitability of the question for that type of graph. \textit{LEAF-QA} contains two types of questions: structural and relational. The structural questions test understanding of the individual elements present in the chart image, while the relational questions require a deeper understanding of the relations between different plot elements. The details of the question templates are provided in the Supplementary Material. \textit{LEAF-QA} does not include value-based questions to avoid complexities due to different formats of data values (e.g. pie chart values or line plot series). 

\begin{figure}[htp!]
\centering
\includegraphics[width=0.9\linewidth]{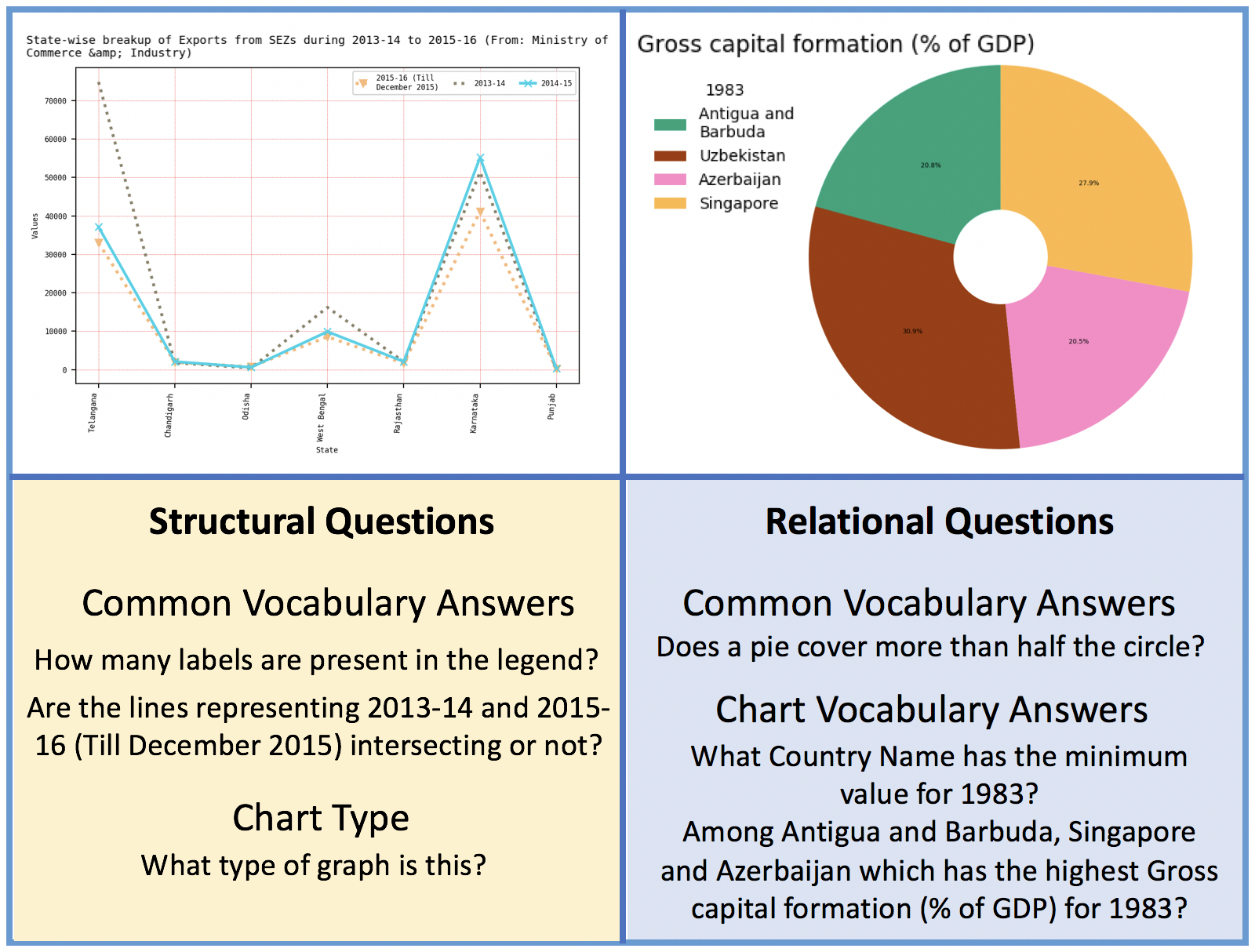}
\caption{Sample question-answers from \textit{LEAF-QA}.}
\label{fig:QuestionAnsSam}
\end{figure}

\textbf{Paraphrases:} For each question template, paraphrases for the question are generated through Google Translate API\footnote{https://cloud.google.com/translate/docs/}, with the number of paraphrases varying from $3-10$ based upon the complexity of the sentence. For each question, one of the paraphrases is randomly selected and the chart specific parts in the paraphrased template are replaced by words from the chart.

\textbf{Answer Types:} The different answer types in \textit{LEAF-QA} are described as follows. The answers which are contained in the chart are referred to as Chart Vocabulary answers. These comprise of answers to questions such as those seeking the label with the maximum value in the chart. A specialized class of answers, namely the Chart Type, deals with questions such as "What type of graph is this?". Finally, the Common Vocabulary answers include common words such as Yes, No, None etc and numbers for questions pertaining to counting. 

\subsection{Training/Test Corpus}
\textit{LEAF-QA} has the following corpus to train/test question answering methods:

\begin{itemize}
\setlength{\itemsep}{0.5pt}
    \item \textbf{Training}: The training set has around $200,000$ chart images, $1.5$ million questions and $2$ million answers. Note that the number of answers is higher because some questions can have multiple answers, e.g. \textit{"Name the years in which US GDP growth rate was higher than in 2004"}. There are around $12,000$ unique answers, which is significantly higher than previous datasets~\cite{ebrahimi2018figureqa,kafle2018dvqa}.
    \item \textbf{Testing}: The testing set has around $40,000$ images, $0.3$ million questions and $0.4$ million answers, generated from the same data sources as training.
    \item \textbf{Novel Testing}: A smaller testing set of $8,150$ images, generated from data sources not seen in the training set\footnote{www.kaggle.com/cdc/tobacco-use,www.kaggle.com/nasa/solar-eclipses,www.kaggle.com/uciml/breast-cancer-wisconsin-data,www.kaggle.com/m4h3ndr4n5p4rk3r/total-number-of-road-kills-in-india-from-20092015}, is provided to test the generalization of question answering models.
\end{itemize}

The detailed break-up of charts and the question and answer types in each corpus is shown in Table~\ref{tab:CorpusBreakUp}.

\begin{figure}[htp!]
    \centering
    \includegraphics[width=1.00\linewidth]{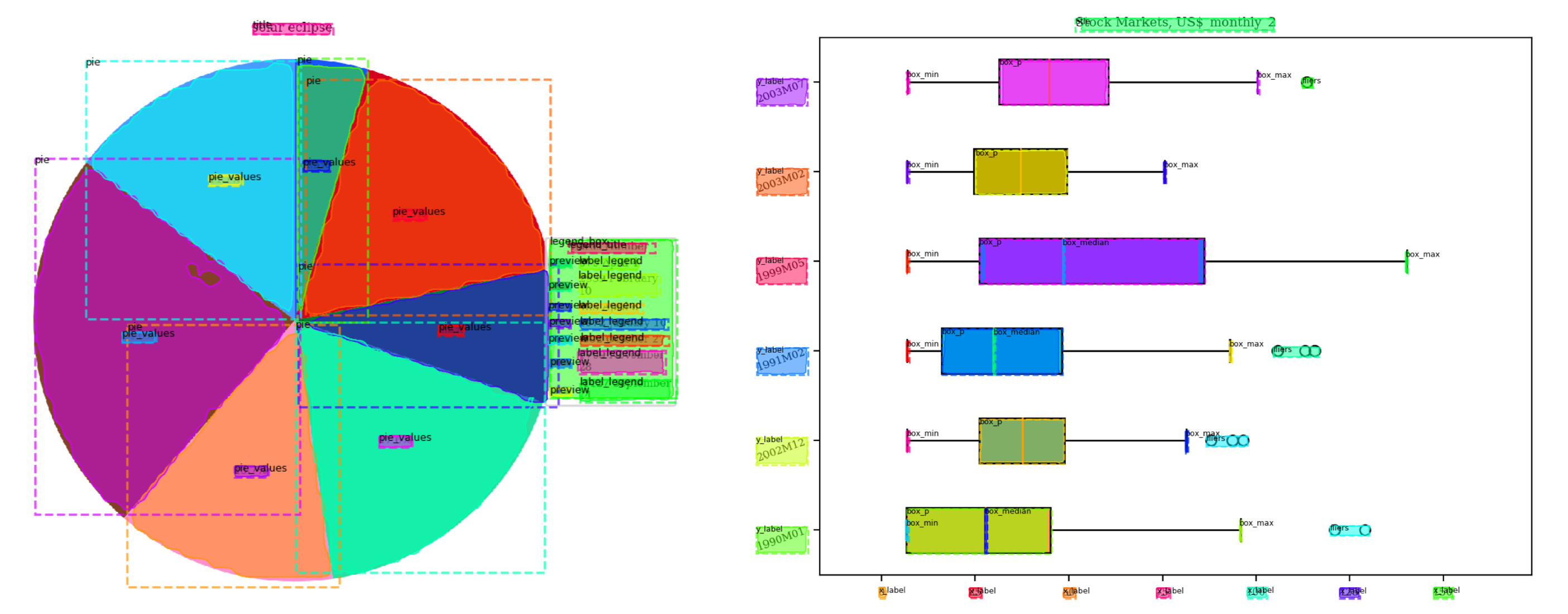}
    \caption{Sample Detection output of the trained Mask R-CNN on charts from the test and novel test sets}
    \label{fig:Detections}
\end{figure}

\begin{figure*}[h]
\centering
\includegraphics[width=0.80\linewidth]{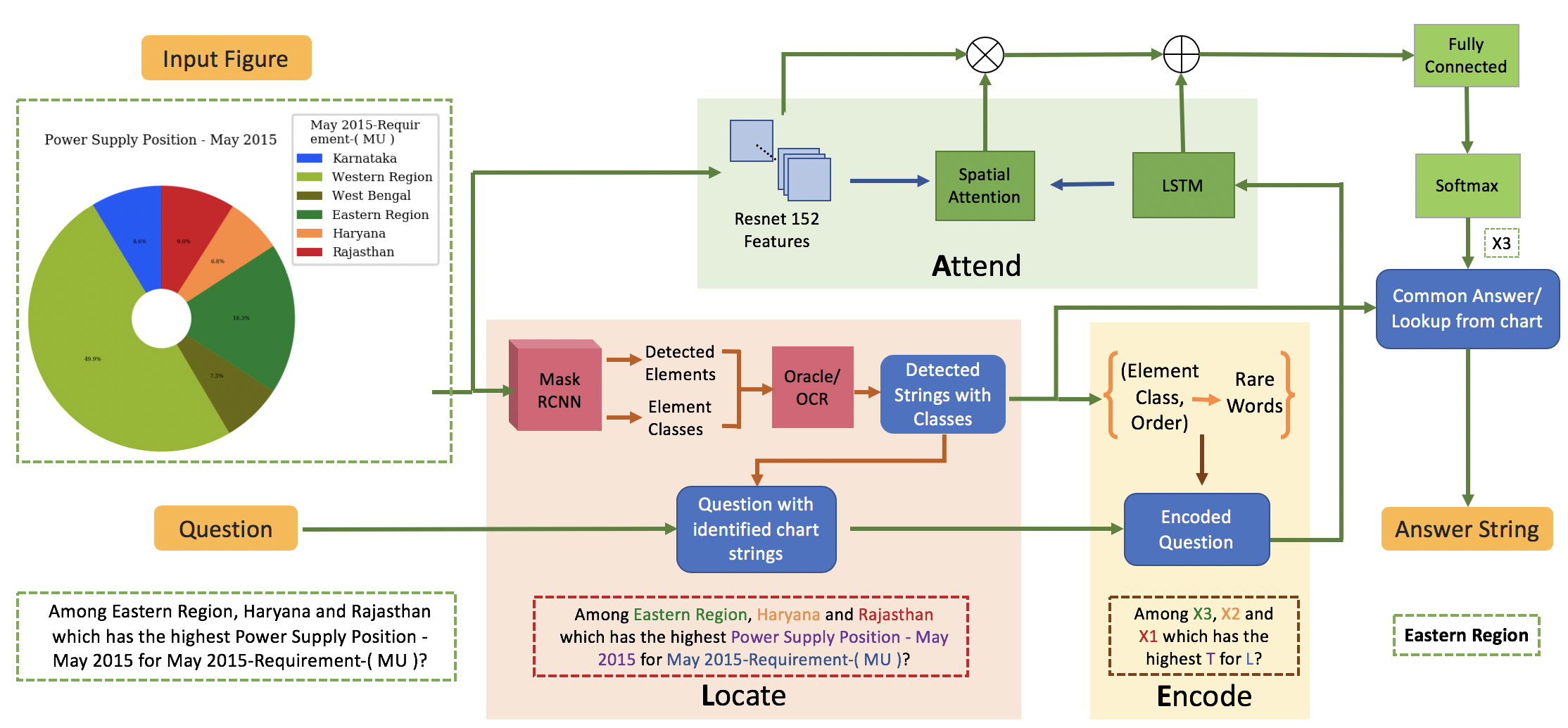}
\caption{The proposed \textit{LEAF-Net} architecture for learning to answer questions on \textit{LEAF-QA}. The question is encoded in terms of the chart elements which are located using the the Mask R-CNN and the Oracle/OCR. This encoded question and the chart image is used to calculate the spatial attention and the final answer, encoded in terms of the detected chart elements.}
\label{fig:OverallArchitecture}
\end{figure*}

\section{\textit{LEAF-Net} baseline}\label{sec:baseline}
In this section, a novel architecture, \textit{LEAF-Net} (shown in Figure~\ref{fig:OverallArchitecture}) is described to address the challenges of question answering on \textit{LEAF-QA}. While visual question answering is a challenging task, \textit{LEAF-QA} has a further challenge to enable multimodal question answering with no fixed vocabulary. \textit{LEAF-Net} enables this through an integrated pipeline consisting of three stages - chart parsing, question and answer encoding and attention-based learning. Each of these is described in details below: 

\subsection{Chart Parsing} \label{sec:MaskRCNN}
A Mask-RCNN network~\cite{he2017mask} using a Resnet-101 backbone has been used to detect and classify elements of the provided chart images. The Matterport implementation\footnote{\text{www.github.com/matterport/Mask\_RCNN}} is used for the network. The training corpus of $200,000$ images along with the bounding boxes/masks and class labels are fed to the network for optimization. The training is done with a batch size of $2$ over $2$ NVIDA K80 GPUs, with a learning rate of $0.001$, for over $1000$ epochs. Some sample detection results from the trained model on the test and novel test sets are shown in Figure~\ref{fig:Detections}. Detailed evaluation is discussed in Section~\ref{sec:Evaluation}.

\subsection{Question-Answer Encoding} \label{sec:TextEncode}
As the question and answers in \textit{LEAF-QA} contain words from chart specific vocabularies, it is necessary to encode them in a consistent way to enable learning by question answering algorithms. For this the following steps are taken:

\textbf{Chart Text Extraction:} Parsed chart components from the previous step, corresponding to text elements, are processed to extract the string inside them. During the training phase, text string from the ground truth box with the maximum overlap (min IOU 0.5) is attached to the detected box. During inference, text strings are extracted using an oracle (assuming ground truth is available) or fully automated using OCR\footnote{https://opensource.google.com/projects/tesseract}. \textit{LEAF-Net} has been tested with both oracle and OCR baselines.

\textbf{Text string ordering}: Each text string is assigned a tuple \textit{(Element Type, Element Order)} where element type is the detected class of the text string. Also, to maintain consistency in value axis of horizontal plots, the element types for x-label and y-label strings are reversed. This is determined based on variation in the horizontal and vertical lengths of the detected bars/boxes. Ordering of each text element type is done as follows: x-labels are sorted left to right and y-labels from bottom to top. The detected legend text elements are sorted top-to-bottom, left-to-right. Pie labels and pie values are sorted clockwise. 

\textbf{Question Encoding}: Questions for a chart are then encoded in terms of the text strings which are extracted before. To identify the correspondence, the question string is matched with the list of extracted text strings, longest first. After each positive match, the matched part of the question is replaced by a specific rare word in the GloVe dictionary~\cite{pennington2014glove}. The rare word is pre-defined in terms of  one-to-one mapping from the tuple \textit{(Element Type, Element Order)} corresponding to the matched text string. The process of obtaining the ordering is detailed in the previous paragraph. The encoded question string is in turn fed to the LSTM network. 

\textbf{Answer Encoding}: The answer text is encoded into a fixed $75$ dimensional vector, with one dimension each for chart text elements, chart types, yes/no/none/negative/positive and numeric values (1-15 in this case). The text elements are ordered similarly as for the question, but using the actual ground truth boxes and annotations, for the purpose of training. The answer is encoded as a 1-0 vector, wherein the dimensions representing the answer text are set to 1.

For example in the pie chart in Figure~\ref{fig:QuestionAnsSam}, the question ``Which country has the minimum value for 1983?'' has the string ``1983'', which is the legend title in the chart. The string is replaced with the rare word mapped to the legend title. Similarly the answer, which is ``Antigua and Barbuda'', is the first legend label, and therefore the answer vector has a $1$ in the index reserved for the first legend label.

\subsection{Attention Network}
Stacked attention networks (SANs)~\cite{Kazemi2017,yang2016stacked} have shown strong performance on the task of question answering on natural images. In \textit{LEAF-Net}, the encoded question-answer pairs and chart images are used to train the attention network. The chart images are resized to $448 \times 448$ and pre-processed using Resnet-152\footnote{https://github.com/KaimingHe/deep-residual-networks} to a $14 \times 14 \times 2048$ encoding. This is concatenated with the output of the LSTM layer applied on the question encoding, and fed to spatial attention network, followed by a fully connected network. Finally, the cross-entropy loss between answer vector and the output of fully connected layer is minimized using an Adam optimizer. The network is trained using NVDIA K80 GPUs for $10,000$ epochs with a batch size of $128$, an initial learning rate of $0.001$, a half-life of $5000$ iterations over the training corpus of $200,000$ images and $1.5$ million question-answers.

\subsection{Inference}
At inference time, the chart image is fed to the Mask-RCNN and the output detections are used to encode the question being asked. The encoded question and the image are then fed to the attention network to get the output vector. The answer is taken to be the chart string corresponding to the highest valued dimension in the output vector. Note that even for questions having multiple answers, the current inference produces only one answer. Multiple answer generation would be a future direction.

\section{Evaluation} \label{sec:Evaluation}
A thorough analysis of the performance of \textit{LEAF-Net} along with different baselines for \textit{LEAF-QA} are described in this section.

\subsection{Chart Parsing Evaluation}
\textbf{Metric Calculation}: The detection performance is computed in terms of precision and recall for each chart element type, averaged over all the charts. Following prior art~\cite{he2017mask, ren2015faster}, we start with the predicted mask with the highest confidence score and the closest unassigned ground truth mask is matched if the mask IoU is greater than $0.5$. The precision is then taken as the ratio of matched predicted masks to total predicted masks, and the recall is taken as the ratio of number of matched predictions to the number of ground truth masks, and then averaged over charts.

\textbf{Analysis}: The detection performance for various text elements is reported in Table~\ref{tab:MRCNNClasswisePrecision}. Analysis of the detection performance of different element types is discussed below:
\begin{itemize}
\setlength{\itemsep}{0.5pt}
    \item \textbf{Title text}: Chart titles and axis titles are detected with high precision for both test and novel test. Chart titles show high recall.
    \item \textbf{Axis Labels}: Axis labels are detected with high precision and recall across both the sets.
    \item \textbf{Legend}: Legend boxes are detected accurately, while the detection scores for legend title, label and preview elements are lower. This reflects the variation in legends across charts.
    \item \textbf{Pies}: While pie values and labels have high recall, precision is lower. This may be because of the varied positioning and sizes of these elements.
\end{itemize}

\subsection{Question Answering Evaluation}

\textbf{Evaluation Metric:} The Question Answering performance is reported in terms of the accuracy of predicting an answer string present in the set of ground truth answers, given that there can be multiple correct answers. For the current baseline, getting one right answer out of multiple possible answers is considered a success. The performance has been broken down by the type of questions and further by the type of answers that the question seeks, and has been shown in Table~\ref{tab:LEAF-QAper}.

\textbf{Baselines:} The following baselines are reported for evaluation:
 \begin{itemize}
\setlength{\itemsep}{0.5pt}
    \item \textbf{QUES:} For QUES baseline, the question string is passed through an LSTM layer and the loss is optimized with the answer vector. This does not take the chart image as input.
    \item \textbf{IMG:} The chart image is re-sized and passed through SAN, with the loss being optimized as before. This assumes no question input.
    \item \textbf{QUES-ENC:} For QUES-ENC, the question string is encoded using the output of chart parsing, which is then passed through an LSTM and optimized with the answer vector. This inherits the parsing structure from the chart image, but the further attention network is not utilized.
    \item \textbf{LEAF-Net:} For \textit{LEAF-Net}, results using both OCR and Oracle are provided.
\end{itemize}
\textbf{Analysis:} The performance break-down has been shown in Table~\ref{tab:LEAF-QAper}. It can be seen that \textbf{QUES}, \textbf{QUES-Enc} and \textbf{IMG} baselines do not generalize well across question sets, however \textit{LEAF-Net} gives a consistent performance. This ablation study shows that \textit{LEAF-QA} answers are not biased, and also highlights the need for robust methods taking both the chart image and question into account to generalize well. There is a drop in performance of \textit{LEAF-Net} on the novel test set as compared to the test set for certain question types. While this highlights the difficulty of the novel test set, it also calls for future work on studying adaptation. A detailed analysis of \textit{LEAF-Net} performance is discussed below:
\begin{itemize}
\setlength{\itemsep}{0.5pt}
    \item \textbf{Structural Questions}: Questions pertaining to chart type have high accuracy across test and novel test for \textit{LEAF-Net}. For common vocabulary answers, there is a fall in accuracy on the novel test, however both Oracle and OCR show similar performance. 
    
    \item \textbf{Relational Questions}: Relational questions with common vocabulary answers show consistent performance for both Oracle and OCR. However, there is an overall decrease in performance for more complex relational questions compared to structural ones. For chart vocabulary answers, there is a fall in accuracy for the OCR case, as it involves parsing longer chart labels.
\end{itemize}

\begin{table*}[htp!]
    \centering
    \captionsetup{skip=0pt}
    \small
    \begin{tabular}{|c|cc|cc|c|cc|cc|}
        \hline
        \multirow{2}{*}{\textbf{Element}} &
        \multicolumn{2}{c|}{\textbf{Test}} &
        \multicolumn{2}{c|}{\textbf{Novel Test}} &
        \multirow{2}{*}{\textbf{Element}} &
        \multicolumn{2}{c|}{\textbf{Test}} &
        \multicolumn{2}{c|}{\textbf{Novel Test}}
        \\\cline{2-3}\cline{4-5}\cline{7-8}\cline{9-10}
         & \textbf{P} &  \textbf{R} & \textbf{P} & \textbf{R} & & \textbf{P} & \textbf{R} & \textbf{P} &  \textbf{R}\\
        \hline
        Title & 0.99 & 0.99 & 0.99 & 0.99 & Legend Box & 0.97 & 0.99 & 0.95 & 0.99 \\
        X-Axis Title & 0.99 & 0.63 & 0.99 & 0.68 & Legend Title & 0.79 & 0.90 & 0.49 & 0.30 \\
        Y-Axis Title & 0.99 & 0.65 & 0.99 & 0.68 & Legend Label & 0.91 & 0.99 & 0.85 & 0.99 \\
        X-Axis Label & 0.99 & 0.98 & 1.00 & 0.99 & Legend Preview & 0.84 & 0.66 & 0.82 & 0.68 \\
        Y-Axis Label & 0.99 & 0.98 & 0.99 & 0.99 & Pie Values & 0.81 & 0.99 & 0.79 & 0.99 \\
        Pie Label & 0.51 & 0.99 & 0.47 & 0.99 & & & & & \\
        \hline
    \end{tabular}
    \vspace{1ex}
    \caption{Class-wise precision (P) and recall (R) scores for text elements in test and novel test charts using Mask-RCNN.}
    \label{tab:MRCNNClasswisePrecision}
\end{table*}

\begin{table*}[htp!]
    \centering
    \captionsetup{skip=0pt}
    \small
    \begin{tabular}{|c|cc|cc|c|}
        \hline
      \multirow{2}{*}{\textbf{Baselines}} &
        \multicolumn{2}{c|}{\textbf{Structural}} &
        \multicolumn{2}{c|}{\textbf{Relational}} & \multirow{2}{*}{\textbf{Overall}}
        \\\cline{2-5}
  
         & Common Vocab Ans & Chart Type & Common Vocab Ans & Chart Vocab Ans & \\
         \hline
         \multicolumn{6}{|c|}{\textbf{Test Set}}  \\
         \hline
         QUES & 55.21 & 21.02 & 50.69 & 0.0 & 18.85 \\
         QUES-ENC (Oracle) & 51.03 & 21.02 & 40.11 & 41.52 & 40.43 \\
         IMG (Oracle) & 4.92 & 0.59 & 13.65 & 30.52 & 22.29 \\
        \textbf{LEAF-Net (Oracle)} & \textbf{90.21} & \textbf{99.89} & \textbf{68.80} & \textbf{60.77} & \textbf{67.42} \\
        \textbf{LEAF-Net (OCR)} & 90.09 & \textbf{99.89} & 65.83 & 42.89 & 55.94 \\
        \hline 
         \multicolumn{6}{|c|}{\textbf{Novel Test Set}}  \\
         \hline
         QUES & 38.49 & 6.26 & 41.94 & 0.0 & 14.18 \\
         QUES-ENC (Oracle) & 39.06 & 6.26 & 43.72 & 39.45 & 38.60 \\
         IMG & 3.52 & 0.0 & 5.72 & 28.55 & 19.05 \\
         \textbf{LEAF-Net (Oracle)} & \textbf{72.28} & \textbf{96.33} & \textbf{62.99} & \textbf{55.78} & \textbf{61.33} \\
         \textbf{LEAF-Net (OCR)} & 72.20 & \textbf{96.33} & 61.25 & 47.26 & 55.73 \\
        \hline
    \end{tabular}
    \vspace{1ex}
    \caption{Question Answering accuracy for different question and answer types on the test and novel test.}
    \label{tab:LEAF-QAper}
\end{table*}

\begin{table*}[htp!]
    \centering
    \captionsetup{skip=0pt}
    \small
    \begin{tabular}{|c|cccc|cccc|}
        \hline
      \multirow{2}{*}{\textbf{Baselines}} &
        \multicolumn{4}{c|}{\textbf{Test-Familiar}} &
        \multicolumn{4}{c|}{\textbf{Test-Novel}} 
        \\\cline{2-9}
  
         & Structure & Data &  Reasoning & \textbf{Overall} & Structure & Data & Reasoning & \textbf{Overall} \\
         \hline
        SAN-VQA\cite{kafle2018dvqa} & 94.71 & 18.78 & 37.29 & 36.04 & 94.82 & 18.92 & 37.25 & 36.14 \\
        MOM\cite{kafle2018dvqa} & 94.71 & 38.20 & 40.99 & 45.03 & 94.82 & 29.14 & 39.26 & 40.90 \\
        SANDY (Oracle)\cite{kafle2018dvqa} & 96.47 & 65.40 & 44.03 & 56.48 & 96.42 &   65.55 & 44.09 & 56.62 \\
        SANDY (OCR)\cite{kafle2018dvqa} & 96.47 & 37.82 & 41.50 & 45.77 & 96.42 & 37.78 & 41.49 & 45.81 \\
        \textbf{LEAF-Net (Oracle)} & \textbf{98.42} & \textbf{81.25} & \textbf{61.38} & \textbf{72.72} & \textbf{98.47} & \textbf{81.32} & \textbf{61.59} & \textbf{72.89} \\
        \textbf{LEAF-Net (OCR)} & \textbf{98.35} & \textbf{74.64} & \textbf{57.96} & \textbf{68.73} & \textbf{98.39} & \textbf{74.11} & \textbf{58.07} & \textbf{68.67} \\
        \hline
    \end{tabular}
    \vspace{1ex}
    \caption{Comparison of Question Answering accuracy on DVQA\cite{kafle2018dvqa} dataset.}
    \label{tab:DVQAper}
\end{table*}

\begin{figure}[htp!]
    \centering
    \includegraphics[width=0.4\textwidth]{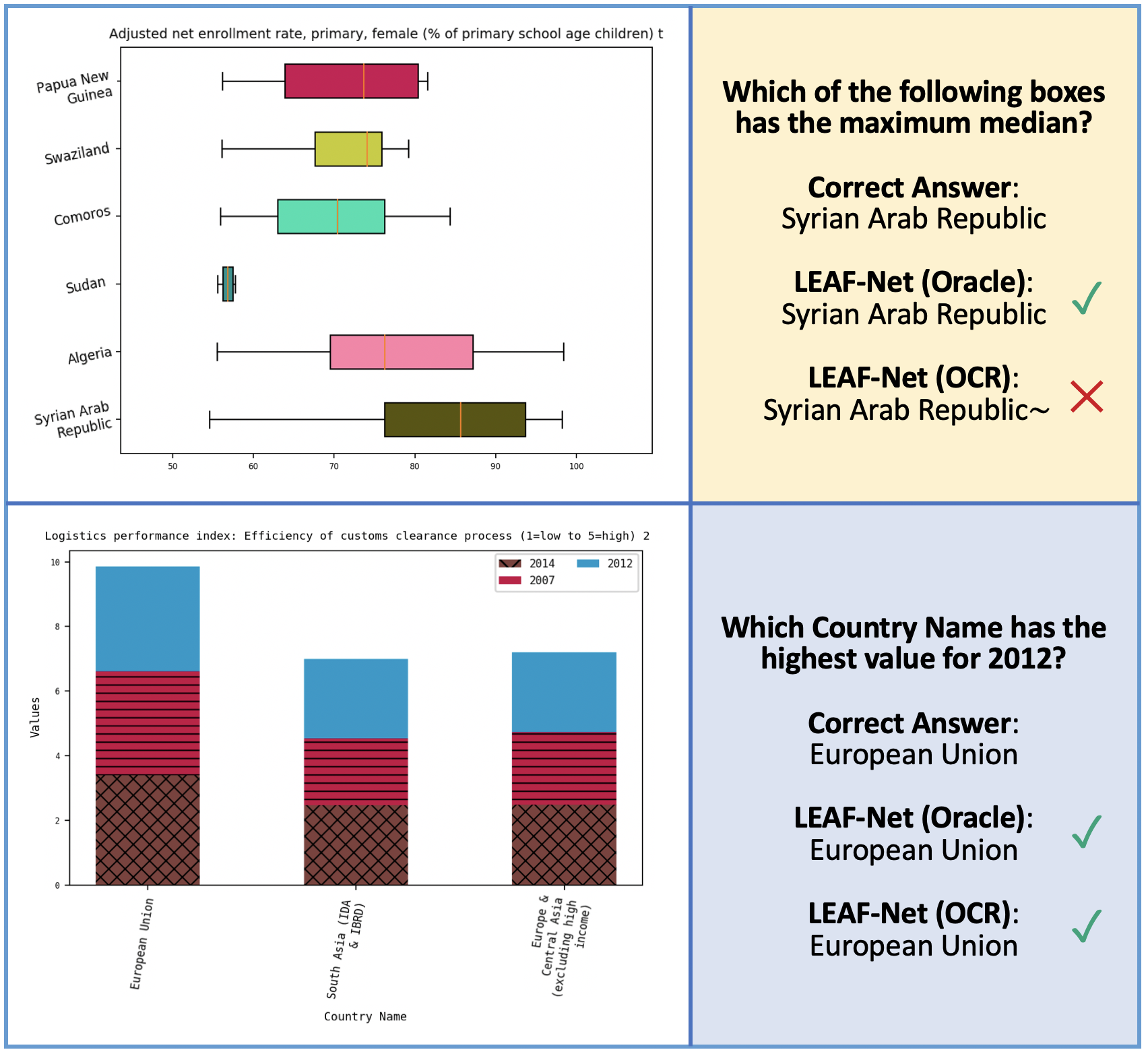}
    \caption{Some examples of predictions using \textit{LEAF-Net (Oracle)} and \textit{LEAF-Net (OCR)}.}
    \label{fig:SampleRuns}
\end{figure}

\textbf{Sample Runs of \textit{LEAF-Net}}: Performance of the model on some example questions from the test set is shown as follows:

In the first chart in Figure~\ref{fig:SampleRuns}, the question seeks the box having the maximum median. Both the Oracle and OCR based models predict the answer index corresponding to the $1^\text{st}$ label as the answer. The Oracle thus predicts the correct string answer but the OCR predicts one extra spurious character and thus makes an incorrect prediction.

In the second example shown in Figure~\ref{fig:SampleRuns}, the question involves comparing the values for a specific data series out of the three series plotted in the chart. Both the networks correctly predict the right index (corresponding to the $1^\text{st}$ Label) as the answer, and also recover the correct strings, thus making correct predictions.

\begin{table}[htp!]
    \centering
    \captionsetup{skip=0pt}
    \begin{tabular}{cc}
       \hline
        \textbf{Method} & \textbf{Accuracy} \\
        \hline
         CNN+LSTM\cite{ebrahimi2018figureqa} & $56.16$ \\
         Relation Net\cite{ebrahimi2018figureqa} & $72.54$ \\
         \textbf{LEAF-Net (OCR)} & $\mathbf{81.15}$ \\
        \hline 
    \end{tabular}
    \vspace{1ex}
    \caption{Validation accuracy on FigureQA~\cite{ebrahimi2018figureqa}.}
    \label{tab:FigureQABase}
\end{table}

\section{Comparisons on DVQA and FigureQA}
To evaluate the performance of \textit{LEAF-Net}, it has been tested on the existing figure question answering datasets, FigureQA\cite{ebrahimi2018figureqa} and DVQA\cite{kafle2018dvqa}. For FigureQA, \textit{LEAF-Net} has been trained with a fixed yes/no answer dictionary, without encoding, since the answers are not contained in the charts. 

For DVQA, \textit{LEAF-Net} has been evaluated against their enlisted baselines, namely SAN, MOM and SANDY. The Mask-RCNN, trained on \textit{LEAF-QA}, has been fine-tuned with the ground truth boxes of DVQA's bar graphs, and the question-answer pairs have been encoded with the chart specific strings. 

As shown in Tables~\ref{tab:DVQAper},~\ref{tab:FigureQABase}, \textit{LEAF-Net} significantly outperforms the prior work on both the datasets. A detailed analysis of comparison with SANDY \cite{kafle2018dvqa} is as below:

\textbf{Overall performance:} \textit{LEAF-Net} significantly outperforms SANDY on all question types, particularly on complex data and reasoning questions. This is due to the consistent chart element-wise encoding in \textit{LEAF-Net} which ensures generalizability on charts with varied vocabularies.

\textbf{OCR:} There is only a slight drop in OCR performance from the oracle performance in \textit{LEAF-Net} as compared to a drastic reduction in SANDY. This is because chart elements are being precisely localized in \textit{LEAF-Net} which leads to improved OCR, as compared to running OCR on the entire chart. This is also evident from the significantly better performance of \textit{LEAF-Net} (OCR) than SANDY (Oracle).

\section{Conclusion and Future Direction}
This paper presents \textit{LEAF-QA}, a comprehensive dataset of charts of varied types constructed from different open real-world sources, along with question answer pairs. The charts have been randomized for visual aspects and the questions have been paraphrased to avoid models from memorizing templates. The dataset is augmented with a novel test set, constructed from unseen data sources to test the generalizability of question answering models. The charts are complemented by dense annotations including masks for each plot element. A strong baseline, \textit{LEAF-Net} comprising of chart parsing, question and answer encoding in terms of chart elements followed by an attention network is proposed. Ablation studies show that \textit{LEAF-Net} performs high consistently across different question types. Further, \textit{LEAF-Net} advances start-of-the-art on DVQA and FigureQA datasets.

\textit{LEAF-QA} is an advancement towards reasoning from figures and charts emulating real-world data. However, there is still significant scope for future work. Firstly, it would be pertinent to study how users query charts/figures and incorporate it into the current question-answer framework. Further, \textit{LEAF-QA} presents a complex reasoning task, and would require exploration of novel deep learning architectures to advance the field. Lastly, it would be useful to incorporate evaluation metrics from information retrieval literature to enable a more holistic view of the question answering performance.

{\small
\bibliographystyle{ieee}
\bibliography{LEAF-QA}
}

\end{document}